\documentclass[journal]{IEEEtran}
\ifCLASSINFOpdf
\else
\fi
\usepackage[hidelinks]{hyperref}   % to hide link borders
\usepackage{hyperref}
\usepackage{subcaption}
\usepackage{epstopdf}
\usepackage{graphicx}
\usepackage{amsfonts}
\usepackage{enumerate}
\usepackage{varioref}
\usepackage{longtable}
\usepackage{graphicx}
\usepackage{amssymb}
\usepackage{chngpage}
\usepackage{multirow}
\usepackage{multirow}
\usepackage{amsthm}
\usepackage{amsmath}
\usepackage{caption}
\usepackage{nccmath}  % to left align an equation
\usepackage[table]{xcolor}  % for color in table
\usepackage{algorithm}% http://ctan.org/pkg/algorithms               ----> % for algorithm (pseudo-code)
\usepackage{algpseudocode}% http://ctan.org/pkg/algorithmicx 		 ----> % for algorithm (pseudo-code)

\usepackage{threeparttable}

\begin{document}

\title{A Fusion-based Gender Recognition Method \\ Using Facial Images}

\author{Benyamin Ghojogh,
		Saeed Bagheri Shouraki,
		Hoda Mohammadzade*,
		Ensieh Iranmehr

\thanks{*Corresponding Author's e-mail: hoda@sharif.edu}
\thanks{All authors are with Department of Electrical Engineering, Sharif University of Technology.}}

%\markboth{To be Submitted to IEEE Transactions on Systems, Man, and Cybernetics - Part C: Applications and Reviews}%

\maketitle

\begin{abstract}
This paper proposes a fusion-based gender recognition method which uses facial images as input. Firstly, this paper utilizes pre-processing and a landmark detection method in order to find the important landmarks of faces. Thereafter, four different frameworks are proposed which are inspired by state-of-the-art gender recognition systems. The first framework extracts features using Local Binary Pattern (LBP) and Principal Component Analysis (PCA) and uses back propagation neural network. The second framework uses Gabor filters, PCA, and kernel Support Vector Machine (SVM). The third framework uses lower part of faces as input and classifies them using kernel SVM. The fourth framework uses Linear Discriminant Analysis (LDA) in order to classify the side outline landmarks of faces. Finally, the four decisions of frameworks are fused using weighted voting.
This paper takes advantage of both texture and geometrical information, the two dominant types of information in facial gender recognition.
Experimental results show the power and effectiveness of the proposed method. This method obtains recognition rate of 94\% for neutral faces of FEI face dataset, which is equal to state-of-the-art rate for this dataset.
\end{abstract}


\IEEEpeerreviewmaketitle
%%%%%%%%%%%%%%%%%%%%%%%%%%%%%%%%%%%%%%%%%
\section{Introduction} 

\IEEEPARstart{G}{ender} recognition can be implemented using different types of data, such as facial images, hand skin images, body signals, and etc. However, as it is the easiest way for human to recognize gender by looking at face, facial images are the most informative data for gender recognition. Since previous decades, various methods and algorithms have been proposed for facial gender recognition. A survey on methods in gender recognition using facial images can be found in \cite{1}.

According to \cite{2,3}, facial information for gender recognition can be categorized into two main categories: (I) geometrical information and (II) texture (or appearance) information. Geometrical information contains the shape and geometry of face and skull, and texture information takes into account the intensity and pattern of facial pixels. Note that in literature of facial gender recognition, usually geometrical information is used in fusion with texture information and not alone; although, texture information is used alone in lots of cases. Moreover, texture and intensity are usually considered the same, referred to as holistic features; however, in \cite{4} intensity, texture, and shape pieces of information are considered differently and are fused in a gender recognition system.

Ref. \cite{5} is one of the first works in facial gender recognition which uses neural network for classification. Neural networks are widely used for feature extraction and classification in gender recognition. For example, back propagation neural networks are used in \cite{3,6,7,8,9,10}. In addition, it is recently shown that convolutional neural networks are useful to obtain discriminative features and classifying genders \cite{11,12,13}.

Other classification methods are also widely used in facial gender recognition, such as Support Vector Machine (SVM), Linear Discriminant Analysis (LDA), and Adaboost. For instance, Independent Component Analysis (ICA) is used in \cite{14} and three different classifiers, i.e., Cosine, LDA, and SVM, are tested. They show that SVM works better with ICA in facial gender recognition. SVM is thoroughly analyzed in \cite{15} for facial gender recognition. In \cite{16}, SVM is used for classifying features extracted by Local Binary Pattern (LBP) and selected by Adaboost. Likewise, SVM is used with LBP in \cite{17}. They also propose a new method in SVM, namely SVM-dropout for avoiding overfitting. Similar to \cite{16}, Adaboost is used with SVM in \cite{18}. Ref. \cite{19,20} are other examples of using SVM in facial gender recognition.

\begin{figure}[!t]
\centering
\includegraphics[width=3.45in]{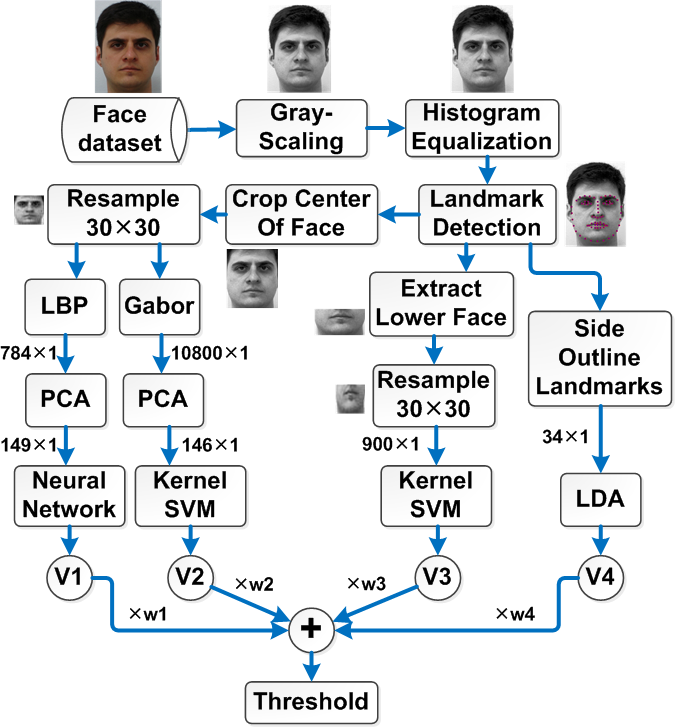}
\caption{Overall structure of the proposed method.}
\label{figure_overall_structure}
\end{figure}

On the other hand, dimension of facial images is usually high, and methods such as Principal Component Analysis (PCA), ICA, or LDA can be used in order to decrease the dimension of data.
In ref. \cite{10,21,22}, ref. \cite{14}, and ref. \cite{23}, PCA, ICA, and LDA are respectively used for feature extraction and dimension reduction.
LDA prepares a proper subspace for projection which results in better discrimination and also dimension reduction. A novel approach is proposed in \cite{24} which randomly crops a set of image patches out of central face and represents each image set as a projection subspace. 

This paper proposes a fusion-based gender recognition method which uses the important ideas of previous work in facial gender recognition area. The overall structure of the proposed method is shown in Fig \ref{figure_overall_structure}. 
As can be seen in this figure, the facial images are firstly pre-processed, i.e., become gray-scaled and histogram equalized. Thereafter, landmarks of faces are detected using one of the landmark detection methods in literature. The center of faces are cropped out of images afterwards. Four frameworks are proposed in this method, inspired by the previous works. The two first use texture information of faces for recognition. The third one uses both geometrical and texture information, and the last one uses merely the geometrical information. The inspiration and reasons are explained for each of these frameworks in the next sections. Finally, decisions of four frameworks are voted with appropriate weights and using a threshold, the final decision on the gender of each test face is obtained.
The remainder of this paper is as follows. Section \ref{section_methodology} introduces the method with details on pre-processing, the four frameworks, and the final fusion. Experimental result is reported in \ref{section_experiment}. The article is finally concluded in Section \ref{section_conclusion}.

\section{Methodology}\label{section_methodology}

The proposed fusion-based facial gender recognition method consists of a pre-processing step, and four frameworks whose decisions are finally fused by weighted voting. In the following, all steps are explained in detail, and their inspiration by previous work are also mentioned.

\subsection{Pre-processing}

As shown in Fig. \ref{figure_overall_structure}, pre-processing phase includes making images gray-scaled, equalizing image histogram, detecting landmarks of faces, cropping center of faces out of images, and finally resampling faces to a $30 \times 30$ uniform grid.

\subsubsection{Gray-scale and Histogram Equalization}
After gray-scaling RGB facial images, they should be histogram equalized for better contrast \cite{25}. This is important in gender recognition because facial details carry discriminative and important information for gender classification. Many other gender recognition methods, such as \cite{23}, also use histogram equalization.

\subsubsection{Landmark Detection}\label{section_landmark}
\begin{figure}[!t]
\centering
\includegraphics[width=3.45in]{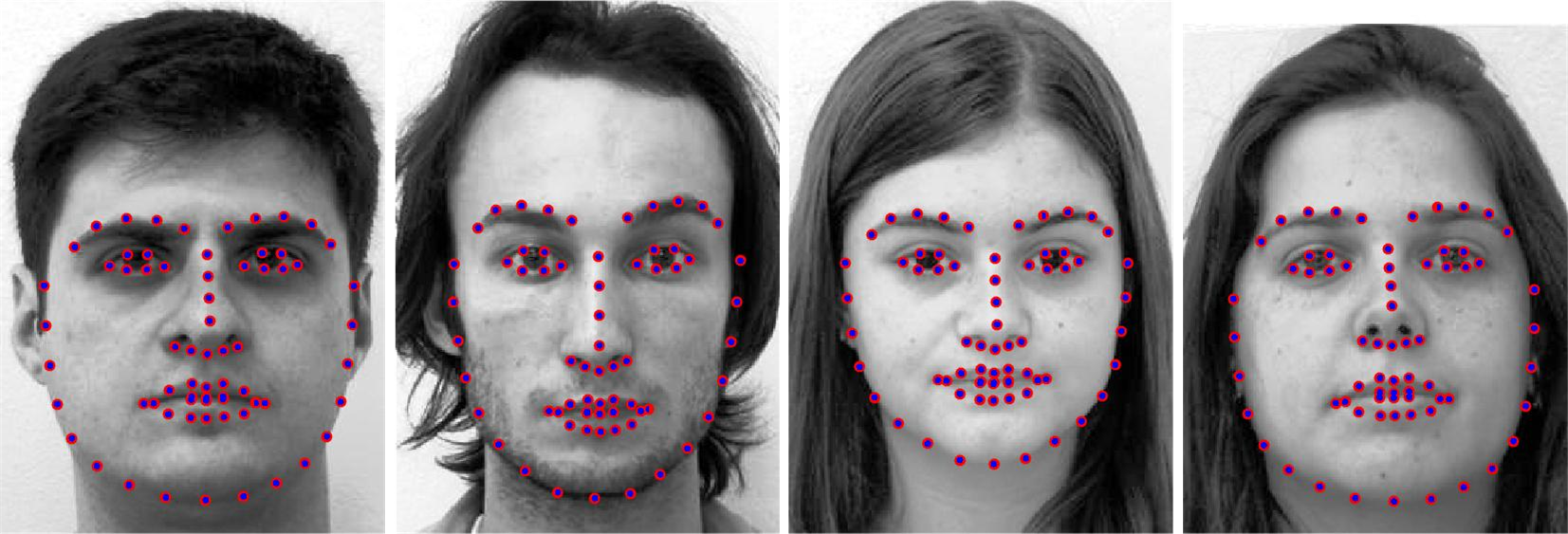}
\caption{Detected landmarks for several face samples.}
\label{figure_landmarks}
\end{figure}

Afterwards, landmarks of faces are detected using a landmark detection method, e.g., Active Shape Model (ASM) \cite{26} or Constrained Local Neural Fields (CLNF) \cite{27}. In this work, CLNF method \cite{27} is used.\footnote{The code of CLNF method can be found in https://github.com/TadasBaltrusaitis/OpenFace.}
In this work, there are 17 landmarks for side outline, 14 landmarks for lips, three landmarks for each upper and lower teeth, six landmarks for each eye, nine landmarks for the whole nose and five landmarks for each eyebrow, resulting in 68 total landmarks. 
These landmarks are required for both geometrical information and cropping issues. 
The detected landmarks on several sample faces are shown in Fig. \ref{figure_landmarks}.

\subsubsection{Cropping Center of Face and Resampling}
In facial gender recognition, solely central part of face should be considered in order to prevent taking advantage of side information such as long hair. Hence, center-of-faces are cropped out of facial images. Different methods are used for face detection and segmentation in literature. For example, Viola-Jones face detector \cite{28} is used in \cite{6,7,10,17}. Some papers have used skin color detection for segmenting central face, such as \cite{3,18,21}. In this work, however, central face is cropped using the side outline landmarks of face.

As the final step of pre-processing, the cropped central faces are resampled to a $30 \times 30$ uniform grid because of two facts: (I) The cropped central faces do not necessarily have the same size in different images, and (II) high number of pixels results in very high dimensional feature vector which is not suitable. 
Moreover, note that not only is not resampling to small size and low resolution images a bad idea, but also as it is declared in \cite{8,15}, information of gender does exist in low resolution of facial images because of skull shape and shadow. In fact, gender can be identified using merely shape and characteristics of skull and face structure.

\subsection{First Framework}

As is shown in Fig. \ref{figure_overall_structure}, the first framework includes uniform LBP, PCA, and back-propagation neural network which are explained in the following.

\subsubsection{Uniform Local Binary Pattern}
It is shown in methods \cite{16,17} that LBP can extract proper features in facial gender recognition. Therefore, in the first framework, inspired by \cite{16,17}, uniform LBP is used for feature extraction out of the $30 \times 30$ face images (see Fig. \ref{figure_overall_structure}). According to \cite{29,30}, LBP for a pixel with intensity $g_c$ and $P$ neighbor pixels with intensities $g_p$ $(p=\{0,\dots,P-1\})$ is calculated as
\begin{equation}
\text{LBP}^u = \sum_{p=0}^{P-1} s(g_p - g_c)2^p,
\end{equation}
where,
\begin{equation}
s(x) =
  \left\{
      \begin{array}{l}
        1 \quad \text{if } x \geq 0,\\
        0 \quad \text{if } x < 0.\\
      \end{array}
    \right.
\end{equation}
If $U(\text{LBP})$ returns the number of transitions from 0 to 1 and vice versa, uniform LBP is calculated as
\begin{equation}
\text{LBP}^u =
  \left\{
      \begin{array}{l}
        \sum_{p=0}^{P-1} s(g_p - g_c) \quad \text{if } U(\text{LBP}) \leq 2,\\
        P+1 \quad \quad \quad \quad \quad \quad \text{otherwise}.\\
      \end{array}
    \right.
\end{equation}
It is shown in \cite{29,30} that uniform LBP works better than LBP; therefore, uniform LBP is used in this framework with eight neighbors having one pixel radius and sliding step of one pixel.

\subsubsection{Principal Component Analysis}\label{section_PCA}

As previously mentioned, because of high dimensionality of image data, PCA can be used to reduce the dimensions. For instance, PCA is used in \cite{10,21,22,23} for feature extraction and reduction. Inspired by them, PCA is used for the sake of dimension reduction. If there are $n$ training samples, the $\text{LBP}^u$ for every training sample $i$ is reshaped to a vector $x_i$, and the covariance matrix of 
\begin{equation}
C = \sum_{i=1}^n (x_i - \overline{x}) (x_i - \overline{x})^{T},
\end{equation}
where $\overline{x}$ is the average of training samples. Afterwards, the eigenvectors ($P_k$) of the covariance matrix are found as
\begin{equation}
C P_k = \lambda_k P_k,
\end{equation}
where $\lambda_k$ denote the eigenvalues of $C$. The eigenvectors whose corresponding eigenvalues are bigger than 0.999 times summation of eigenvalues are kept and the others are left over. In test phase, the reshaped $\text{LBP}^u$ is projected onto the purified eigenvectors and its dimension becomes much lower.

\subsubsection{Neural Network}

As previously mentioned, back-propagation \cite{3,6,7,8,9,10} and convolutional neural networks \cite{11,12,13} are widely used in facial gender recognition, and they are shown to be effective for this application. In this framework, a back-propagation neural network with 149 input neurons, one hidden layer with 10 neurons, and one output neuron is utilized. Labels $1$ and $-1$ are used respectively for females and males, as well as for classifiers of other three frameworks.

\subsection{Second Framework}

As is shown in Fig. \ref{figure_overall_structure}, the second framework consists of Gabor filtering, PCA, and kernel SVM which are detailed in the following.

\subsubsection{Gabor Filtering}

\begin{figure}[!t]
\centering
\includegraphics[width=3.45in]{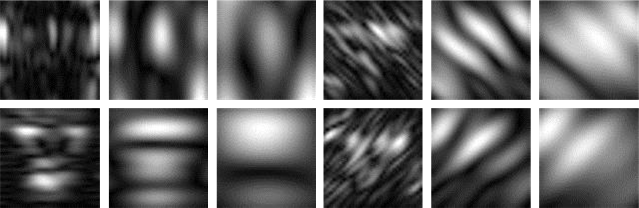}
\caption{Magnitude domain of 12 Gabor filter bank results for a sample face.}
\label{figure_gabor}
\end{figure}

It is shown in \cite{6,18} that Gabor filtering prepares suitable features for gender classification. Inspired by them, Gabor filtering is performed in this framework with three scales (wavelengths 2, 5, and 8) and four orientations (angles 0, 45, 90, and 135 degrees). According to \cite{31}, Gabor filter bank is formulated as
\begin{equation}
g(x,y;\lambda,\theta,\phi,\sigma) = \frac{1}{2\pi \sigma^2} e^{\frac{-(x^2+y^2)}{2\sigma^2}} \text{sin}(\frac{2\pi x'}{\lambda} + \phi),
\end{equation}
where 
\begin{equation}
x' = x \text{ cos}\theta + y \text{ sin}\theta,
\end{equation}
and $\lambda$, $theta$, $\phi$, and $\sigma$ are respectively wavelength (scale), orientation angle, phase offset, and Gaussian standard deviation. In this framework, magnitudes of Gabor filter results are used. Figure \ref{figure_gabor} shows the 12 filter bank results for one of the faces.

\subsubsection{Principal Component Analysis}

There are 12 Gabor filters and each of them have $30 \times 30$ size because of image size. Therefore, after concatenating filter results and reshaping them to a vector, dimension of data becomes 10800 which is very high. PCA is again used for dimension reduction the same as explained in Section \ref{section_PCA}. 

\subsubsection{Kernel SVM}\label{section_kernel_svm}

Various gender recognition methods, such as \cite{14,15,16,17,18,19,20}, use SVM as classifier. This fact shows the effectiveness of this powerful classifier in facial gender recognition.
That is why kernel SVM is used as classifier in the second framework.
According to \cite{32}, the decision function in kernel SVM can be expressed as,
\begin{equation}
y_t = \text{sign} (\sum_{i=1}^N \alpha_i y_i K(x_i,x_t) + b), 
\end{equation}
where $x_i$ and $x_t$ are respectively training and testing data, $b$ is bias, $\alpha_i$ is positive Lagrange multiplier, $N$ is the number of all training samples, $y_i \in \{+1,-1\}$ is the label of training data, and $K(.,.)$ is the kernel function. In this work, radial basis function is used for kernel, which is
\begin{equation}
K(x_i,x_t) = \text{exp}(-||x_i-x_t||^2).
\end{equation}

\subsection{Third Framework}

As is shown in Fig. \ref{figure_overall_structure}, the third framework includes extracting lower part of face, resampling the cropped image, and kernel SVM which are explained in the following.

\subsubsection{Extracting lower part of face}

\begin{figure}[!t]
\centering
\includegraphics[width=3.45in]{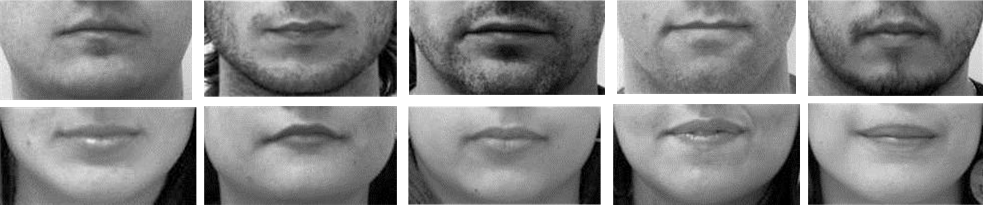}
\caption{Lower part of faces extracted from sample faces. The first and second rows contain male and female samples respectively.}
\label{figure_lower_face}
\end{figure}

According to \cite{19}, lower part of face, i.e., from nose tip to chin, includes crucial information about gender. In other words, gender can be recognized by merely considering the lower part of face because of differences in smoothness or roughness of texture pattern. Inspired by \cite{19}, lower part of face is used in this framework. To extract the lower face, nose tip, chin, and two of the side outline landmarks are used in this work. Several examples of extraction of lower part of face are shown in Fig. \ref{figure_lower_face}. The differences in smoothness of texture in males and females are obvious in this figure.

\subsubsection{Resampling}

In order to decrease dimension of data and also equalize the dimension of data samples, the extracted lower parts of faces are resampled to a uniform $30 \times 30$ grid. Thereafter, the $30 \times 30$ images are reshaped to $900 \times 1$ column vectors.

\subsubsection{Kernel SVM}

Finally, the same as explanations in Section \ref{section_kernel_svm}, kernel SVM is used in order to classify the $900 \times 1$ data samples.

\subsection{Fourth Framework}

As is depicted in Fig. \ref{figure_overall_structure}, the fourth framework includes using LDA to classify the side outline landmarks. It is explained in the following.

\subsubsection{Side Outline Landmarks}
In this framework, the side outline landmarks, which were detected previously (see Section \ref{section_landmark}), are taken into account. The inspirations of this framework are \cite{8,9} in gender recognition and \cite{33,34,35,36} in human biology science, which claim that physical and geometrical characteristics of human face and skull have significant differences in males and females. Side outline of face, eyes, eyebrows, chin, supraorbital ridges, cheekbones, mandible, palate, vertical distance between eyes and nose tip, vertical distance between eyes and lips center, nose width, lips width, height of face, and shadow on face because of facial structure are mentioned in literature as geometrical differences in faces of males and females.
Some of these characteristics might be imperfect because they can be affected by the fatness and slimness of face. According to \cite{8}, side outline shape of faces are the most discriminative feature among geometrical information for goal of gender identification. The above part of side outline, however, is covered by hair and is useless. Hence, the bottom side outline of face from left to right ear is considered in this work. The $x$ and $y$ coordinates of landmarks are all concatenated as a column vector in order to feed the classifier which is explained in the following.

\subsubsection{Linear Discriminant Analysis}

Fisher LDA is used for classification in the fourth framework. LDA is a powerful classifier which projects data onto a discriminative subspace \cite{37}. 
Suppose $S_b$ and $S_w$ denote the between-class and within-class scattering matrices, respectively, formulated as
\begin{equation}
S_b = \sum_{i=1}^C N_i (\mu_i - \mu) (\mu_i - \mu)^T,
\end{equation}
\begin{equation}
S_w = \sum_{i=1}^C \sum_{x_k \in X_i} (x_k - \mu_i) (x_k - \mu_i)^T,
\end{equation}
where $\mu_i$, $\mu$, $N_i$, $x_k$, and $C$ are respectively average of $i^{th}$ class, average of averages of classes, the number of samples of $i^{th}$ class, the $k^{th}$ sample of $i^{th}$ class ($X_i$), and the number of classes which is two here. In this work, $x_k$ is the column vector containing $x$ and $y$ coordinates of side outline landmarks.
The projection discriminative subspace of LDA is constructed by eigenvectors of $S_w^{-1}S_b$.

\subsection{Fusion}

Output labels (decisions) of the four frameworks ($v_i$) are $+1$ for females and $-1$ for males. In order to fuse the four decisions and make the final decision, majority voting can be useful; however, even number of frameworks might result in a draw. Therefore, a better fusion is required, such as weighted voting. The voting weight of every framework can be considered as its recognition rate on a training subset. In other words, weights are better to show the power of their corresponding frameworks. If $w_i$ denotes the voting weight of $i^{th}$ framework, the final decision can be obtained using a simple threshold as
\begin{equation}
\text{Gender} =
  \left\{
      \begin{array}{l}
        \text{Female} \quad \quad \text{if} \quad \sum_{i=1}^4 w_i v_i \geq 0,\\
        \text{Male} \quad \quad \quad \text{if} \quad \sum_{i=1}^4 w_i v_i < 0.\\
      \end{array}
    \right.
\end{equation}
Fusion step is illustrated in Fig. \ref{figure_overall_structure}.

\section{Experimental Result}\label{section_experiment}

\subsection{Utilized dataset}

For verification of the proposed method, experiments were performed on FEI face dataset \cite{38,39}. In this work, solely 200 neutral faces of FEI dataset was used because gender recognition methods, as well as lots of papers in literature, usually deal with neutral faces without pose and expression difference. 100 of faces are pictures of females and 100 faces are males. 75\% of the samples were used as training samples and the rest of samples were considered for test. The number of female and male samples were fairly equal in training and testing phases. 

\subsection{Results}

Table \ref{table_rates} reports the gender recognition rates in the four proposed frameworks and for males, females, and overall. It also reports the rates of final fusion. As can be seen in this table, the best and weakest ones among frameworks are the third and first frameworks respectively.
Moreover, as can be seen in this table, some frameworks work better for males and some are more accurate for females. This, itself, shows that fusion of different frameworks can result in an overall framework which works precisely for both males and females (see last column in Table \ref{table_rates} in which male and female rates are both good and close to each other).
The overall rate in fusion-based method is obtained as 94\% for neutral faces of FEI dataset.

\begin{table}[!t]
%\begin{minipage}{\textwidth}
\renewcommand{\arraystretch}{1.3}  %%% each row size
\caption{Gender recognition rates. F1, F2, F3, and F4 denotes first, second, third, and fourth frameworks respectively. The last row determines the weights associated to frameworks in final fusion.}
\label{table_rates}
\centering
\begin{tabular}{l | c | c | c | c || c}
\hline
\hline
 & F1 & F2 & F3 & F4 & Fusion  \\ 
\hline
Males   & 84\% & 80\% & 100\% & 80\% & 96\%  \\
\hline
Females & 72\% & 92\% & 80\% & 92\% & 92\%  \\
\hline
Overall & 78\% & 86\% & 90\% & 86\% & \textbf{94\%}  \\
\hline
Weight & 0.78 & 0.86 & 0.9 & 0.86 & --  \\
\hline
\hline
\end{tabular}%
%\end{minipage}
\end{table}

\subsection{Comparison With State-of-the-art}

The recognition rates of state-of-the-art methods in neutral faces of FEI dataset are reported in Table \ref{table_comparison} for the sake of comparison. As can be seen in this table, the paper obtains excellent recognition rate (94\%) equal to the state-of-the-art rate \cite{10} reported in literature. According to this table, it is also observed that the recognition rates of males and females are more close to each other in this work in comparison to \cite{3}, as well as being high enough. This fact is obtained because of the advantage of fusing different frameworks and having their benefits together. Every framework extracts specific and different features necessary for gender recognition, and fusing the results brings various types of features and information together for better performance.

\begin{table}[!t]
%\begin{minipage}{\textwidth}
\renewcommand{\arraystretch}{1.3}  %%% each row size
\caption{Comparison of the proposed method with state-of-the-art facial gender recognition methods in FEI dataset. Question marks mean that the paper has not reported the corresponding rates.}
\label{table_comparison}
\centering
\begin{tabular}{l || c | c | c}
\hline
\hline
 & Males & Females & Total \\ 
\hline
\cite{3} & 80\% & \textbf{100\%} & 90\%  \\
\hline
\cite{10} & ? & ? & \textbf{94\%}  \\
\hline
Ours & \textbf{96\%} & 92\% & \textbf{94\%}  \\
\hline
\hline
\end{tabular}%
%\end{minipage}
\end{table}

\section{Conclusion and Discussion}\label{section_conclusion}

In this paper, a fusion-based facial gender recognition method is proposed in which four different frameworks, inspired by the state-of-the-art gender recognition methods, exist. 
The first framework is composed of LBP, PCA, and back propagation neural network, while the second framework included Gabor filtering, PCA, and kernel SVM. The third and fourth frameworks respectively deal with lower part of faces and side outline landmarks, and use kernel SVM and LDA for classification. 

As is obvious, different classification methods, which are neural network, kernel SVM, and LDA, are used in this method making it powerful and effective for different aspects and types of features. In addition, different feature extraction methods, which are LBP, Gabor, PCA, lower face, and landmarks, use almost all the most important features required for gender recognition. 
In this paper, both texture and geometrical information are used for recognizing facial gender. 
The first and second frameworks are using texture information and the third and fourth ones use the geometrical information.
Taking advantage of both types of information and also different frameworks makes the proposed method powerful, as the experiments verify it.

% or


\appendices
%%%%%\section{Proof of the First Zonklar Equation}
%%%%%Appendix one text goes here.

%%%%%\section{}
%%%%%Appendix two text goes here.

%%%%%\section*{Acknowledgment}

%%%%%%%%%%%%%%%The authors would like to thank...

\ifCLASSOPTIONcaptionsoff
  \newpage
\fi



%\bibliographystyle{IEEEtran}


\hfill \break

\end{document}